\documentclass[11pt,letterpaper]{article}
\usepackage[letterpaper]{geometry}
\usepackage{acl2012}
\usepackage{times}
\usepackage{latexsym}
\usepackage{amsmath}
\usepackage{multirow}
\usepackage{url}
\usepackage{tikz}
\usepackage{xcolor}
\usepackage{bm}
\usepackage{lipsum}
\makeatletter
\newcommand{\@BIBLABEL}{\@emptybiblabel}
\newcommand{\@emptybiblabel}[1]{}
\makeatother
\usepackage[hidelinks]{hyperref}

\newcommand\blfootnote[1]{%
  \begingroup
  \renewcommand\thefootnote{}\footnote{#1}%
  \addtocounter{footnote}{-1}%
  \endgroup
}

\title{Neural Lattice Language Models}

\author{Jacob Buckman\footnotemark \\
  Language Technologies Institute \\
  Carnegie Mellon University \\
  {\tt jacobbuckman@cmu.edu} \\\And
  Graham Neubig \\
  Language Technologies Institute \\
  Carnegie Mellon University \\
  {\tt gneubig@cs.cmu.edu} \\}

\date{}

\begin{document}
\maketitle
\blfootnote{* Now at Google: {\tt buckman@google.com}}
\begin{abstract}
  In this work, we propose a new language modeling paradigm that has the ability to perform both prediction and moderation of information flow at multiple granularities: \emph{neural lattice language models}. These models construct a lattice of possible paths through a sentence and marginalize across this lattice to calculate sequence probabilities or optimize parameters. This approach allows us to seamlessly incorporate linguistic intuitions -- including polysemy and existence of multi-word lexical items -- into our language model. Experiments on multiple language modeling tasks show that English neural lattice language models that utilize polysemous embeddings are able to improve perplexity by 9.95\% relative to a word-level baseline, and that a Chinese model that handles multi-character tokens is able to improve perplexity by 20.94\% relative to a character-level baseline.
\end{abstract}

\section{Introduction}

Neural network models have recently contributed towards a great amount of progress in natural language processing. These models typically share a common backbone: recurrent neural networks (RNN), which have proven themselves to be capable of tackling a variety of core natural language processing tasks \cite{hochreiter1997long,elman1990finding}.
One such task is language modeling, in which we estimate a probability distribution over sequences of tokens that corresponds to observed sentences (\S\ref{sec:background}). Neural language models, particularly models conditioned on a particular input, have many applications including in machine translation \cite{bahdanau2016end}, abstractive summarization \cite{chopra2016abstractive}, and speech processing \cite{graves2013speech}. Similarly, state-of-the-art language models are almost universally based on RNNs, particularly long short-term memory (LSTM) networks \cite{jozefowicz2016exploring,inan2016tying,merity2016pointer}.

\begin{figure}
    \centering
    \includegraphics[scale=.15,trim=1.2cm 0 0 0]{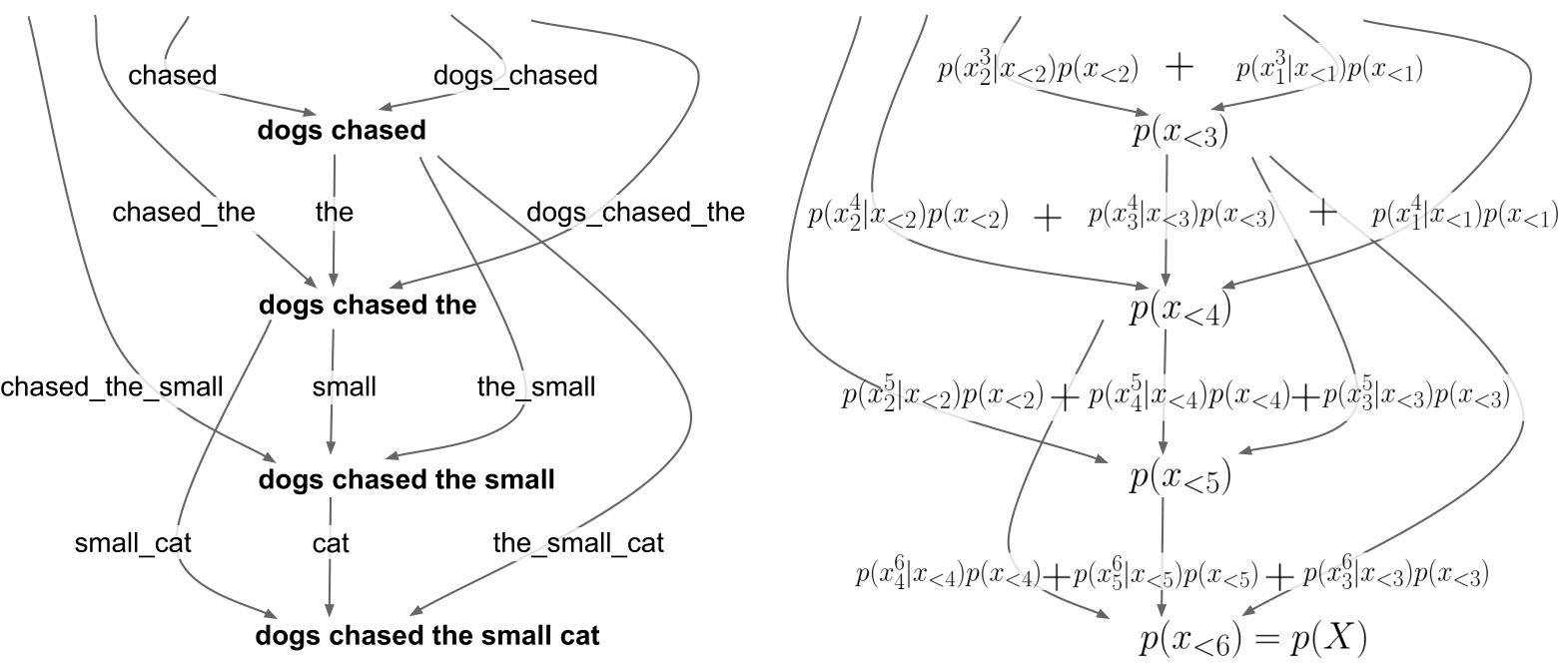}
    \caption{Lattice decomposition of a sentence and its corresponding lattice language model probability calculation}
    \vspace{-2mm}
    \label{fig:latticelm}
\end{figure}

While powerful, LSTM language models usually do not \textit{explicitly} model many commonly-accepted linguistic phenomena. As a result, standard models lack linguistically informed inductive biases, potentially limiting their accuracy, particularly in low-data scenarios \cite{adams2017,koehn}. In this work, we present a novel modification to the standard LSTM language modeling framework that allows us to incorporate some varieties of these linguistic intuitions seamlessly:
\textit{neural lattice language models} (\S\ref{sec:proposed}). Neural lattice language models define a lattice over possible paths through a sentence, and maximize the marginal probability over all paths that lead to generating the reference sentence, as shown in Fig. \ref{fig:latticelm}. Depending on how we define these paths, we can incorporate different assumptions about how language should be modeled.

In the particular instantiations of neural lattice language models covered by this paper, we focus on two properties of language that could potentially be of use in language modeling: the existence of multi-word lexical units \cite{zgusta1967multiword} (\S\ref{sec:multitoken}) and polysemy \cite{ravin2000polysemy} (\S\ref{sec:polysemy}). Neural lattice language models allow the model to incorporate these aspects in an end-to-end fashion by simply adjusting the structure of the underlying lattices.

We run experiments to explore whether these modifications improve the performance of the model (\S\ref{sec:experiments}). Additionally, we provide qualitative visualizations of the model to attempt to understand what types of multi-token phrases and polysemous embeddings have been learned.

\section{Background}
\label{sec:background}

\subsection{Language Models}

Consider a sequence $X$ for which we want to calculate its probability.
Assume we have a vocabulary from which we can select a unique list of $|X|$ tokens $x_1,x_2,\ldots,x_{|X|}$ such that $X = [x_1;x_2;\ldots;x_{|X|}]$, i.e. the concatenation of the tokens (with an appropriate delimiter).
These tokens can be either on the character level \cite{hwang2017character,DBLP:journals/corr/LingTDB15} or word level \cite{inan2016tying,merity2016pointer}.
Using the chain rule, language models generally factorize $p(X)$ in the following way:
\begin{align}
\label{eq:regmarg}
p(X) &= p(x_1,x_2,\ldots,x_{|X|}) \nonumber \\
     &= \prod_{t=1}^{|X|}p(x_t\mid x_1,x_2,\ldots,x_{t-1})
\end{align}

Note that this factorization is exact only in the case where the segmentation is unique.
In character-level models, it is easy to see that this property is maintained, because each token is unique and non-overlapping.
In word-level models, this also holds, because tokens are delimited by spaces, and no word contains a space.

\subsection{Recurrent Neural Networks}

Recurrent neural networks have emerged as the state-of-the-art approach to approximating $p(X)$.
In particular, the LSTM cell \cite{hochreiter1997long} is a specific RNN architecture which has been shown to be effective on many tasks, including language modeling \cite{press2016using,jozefowicz2016exploring,merity2016pointer,inan2016tying}.%
\footnote{In this work, we utilize an LSTM with linked input and forget gates, as proposed by \newcite{greff2016lstm}.}
LSTM language models recursively calculate the hidden and cell states ($h_t$ and $c_t$ respectively) given the input embedding $e_{t-1}$ corresponding to token $x_{t-1}$:
\begin{align}
\label{eqn:lstm}
h_t, c_t = \text{LSTM}(h_{t-1},c_{t-1},e_{t-1},\theta),
\end{align}
then calculate the probability of the next token given the hidden state, generally by performing an affine transform parameterized by $W$ and $b$, followed by a softmax:
\begin{align}
\label{eq:softmax}
p(x_t \mid h_t) := \text{softmax}(W * h_t + b).
\end{align}

%

\section{Neural Lattice Language Models}

\subsection{Language Models with Ambiguous Segmentations}
\label{sec:proposed}

To reiterate, the standard formulation of language modeling in the previous section requires splitting sentence $X$ into a unique set of tokens $x_1,\ldots,x_{|X|}$.
Our proposed method generalizes the previous formulation to remove the requirement of uniqueness of segmentation, similar to that used in non-neural $n$-gram language models such as \newcite{dupont1997lattice} and \newcite{goldwater2007distributional}.

First, we define some terminology.
We use the term ``token'', designated by $x_i$, to describe any indivisible item in our vocabulary that has no other vocabulary item as its constituent part.
We use the term ``chunk'', designated by $k_i$ or $x_i^j$, to describe a sequence of one or more tokens that represents a portion of the full string $X$, containing the unit tokens $x_i$ through $x_j$: $x_i^j = [x_i,x_{i+1};\ldots;x_j]$.
We also refer to the ``token vocabulary'', which is the subset of the vocabulary containing only tokens, and to the ``chunk vocabulary'', which similarly contains all chunks.

Note that we can factorize the probability of any sequence of chunks $K$ using the chain rule, in precisely the same way as sequences of tokens:
\begin{align}
\label{eq:regmarg}
p(K) &= p(k_1,k_2,\ldots,k_{|K|}) \nonumber \\
     &= \prod_{t=1}^{|K|}p(k_t\mid k_1,k_2,\ldots,k_{t-1})
\end{align}

We can factorize the overall probability of a token list $X$ in terms of its chunks by using the chain rule, and marginalizing over all segmentations. 
For any particular token list $X$, we define a set of valid segmentations $\mathcal{S}(X)$, such that for every sequence $s \in \mathcal{S}(X)$, $X = [x_{s_0}^{s_1-1};x_{s_1}^{s_2-1};\ldots;x_{s_{|s|-1}}^{s_{|s|}}]$.
The factorization is:
\small
\begin{align}
\label{eq:latmarg}
p(X) &= \sum_S p(X, S) = \sum_S p(X|S) p(S) = \sum_{S \in \mathcal{S}(X)} p(S) \nonumber \\
     &= \sum_{S \in \mathcal{S}(X)}\prod_{t=1}^{|S|}p(x_{s_{t-1}}^{s_t-1}\mid x_{s_0}^{s_1-1},x_{s_1}^{s_2-1},\ldots,x_{s_{t-2}}^{s_{t-1}-1})
\end{align}
\normalsize

Note that, by definition, there exists a unique segmentation of $X$ such that $x_1,x_2,\ldots$ are all tokens, in which case $|S|=|X|$.
When only that one unique segmentation is allowed per $X$, $\mathcal{S}$ contains only that one element, so summation drops out, and therefore for standard character-level and word-level models, Eq.~(\ref{eq:latmarg}) reduces to Eq.~(\ref{eq:regmarg}), as desired. 
However, for models that license multiple segmentations per $X$, computing this marginalization directly is generally intractable.
For example, consider segmenting a sentence using a vocabulary containing all words and all 2-word expressions.
The size of $\mathcal{S}$ would grow exponentially with the number of words in $X$, meaning we would have to marginalize over trillions of unique segmentations for even modestly-sized sentences.

\subsection{Lattice Language Models}

To avoid this, it is possible to re-organize the computations in a lattice, which allows us to dramatically reduce the number of computations required \cite{dupont1997lattice,neubig2010learning}.

All segmentations of $X$ can be expressed as the edges of paths through a lattice over token-level prefixes of $X$: $x_{<1}, x_{<2}, \ldots, X$. The infimum is the empty prefix $x_{<1}$; the supremum is $X$; an edge from prefix $x_{<i}$ to prefix $x_{<j}$ exists if and only if there exists a chunk $x_i^j$ in our chunk vocabulary such that $[x_{<i};x_i^j] = x_{<j}$.
Each path through the lattice from $x_{<1}$ to $X$ is a segmentation of $X$ into the list of tokens on the traversed edges, as seen in Fig. \ref{fig:latticelm}.

The probability of a specific prefix $p(x_{<j})$ is calculated by marginalizing over all segmentations leading up to $x_{j-1}$
\begin{equation}
  p(x_{<j}) = \sum_{S \in \mathcal{S}(x_{<j})} \prod_{t=1}^{|S|} p(x_{s_{t-1}}^{s_t-1} \mid x_{<s_{t-1}}),
\end{equation}
where by definition $s_{|S|}=j$.
The key insight here that allows us to calculate this efficiently is that this is a recursive formula and that instead of marginalizing over all segmentations, we can marginalize over immediate predecessor edges in the lattice, $A_j$. Each item in $A_j$ is a location $i$ ($=s_{t-1}$), which indicates that the edge between prefix $x_{<i}$ and prefix $x_{<j}$, corresponding to token $x_i^j$, exists in the lattice. We can thus calculate $p(x_{<j})$ as
\begin{align}
\label{eq:neuralsum}
p(x_{<j}) = \sum_{i \in A_j}p(x_{<i})p(x_i^j \mid x_{<i}).
\end{align}

Since $X$ is the supremum prefix node, we can use this formula to calculate $p(X)$ by setting $j = |X|$. In order to do this, we need to calculate the probability of each of its $|X|$ predecessors. Each of those takes up to $|X|$ calculations, meaning that the computation for $p(X)$ can be done in O($|X|^2$) time. If we can guarantee that each node will have a maximum number of incoming edges $D$ so that $|A_j| \leq D$ for all $j$, then this bound can be reduced to O($D|X|$) time.%
\footnote{Thus, the standard token-level language model where $D=1$ takes $O(|X|)$ computations.}

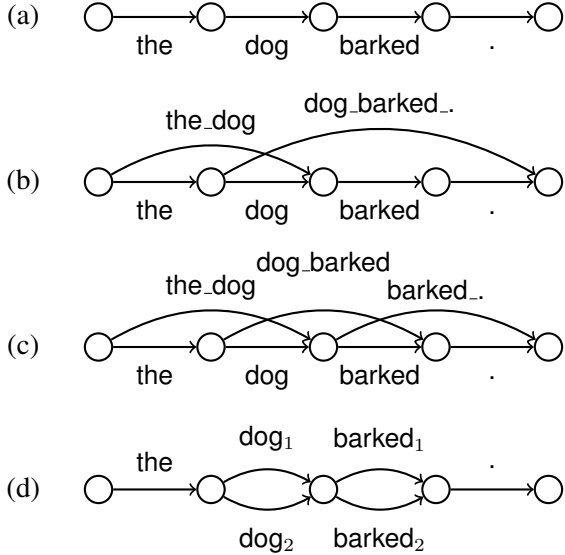
\begin{figure}
    \centering
        \begin{tikzpicture}[->,auto,node distance=1cm,
          thick,main node/.style={circle,draw,font=\sffamily\Large\bfseries}]
        
          \node[main node] (1) {};
          \node[main node] (2) [right of=1, right=.3cm] {};
          \node[main node] (3) [right of=2, right=.3cm] {};
          \node[main node] (4) [right of=3, right=.3cm] {};
          \node[main node] (5) [right of=4, right=.3cm] {};
        
          \path[every node/.style={font=\sffamily\small}]
            (1) edge node [below=.1cm] {the} (2)
            (2) edge node [below=.1cm] {dog} (3)
            (3) edge node [below=.1cm] {barked} (4)
            (4) edge node [below=.25cm] {.} (5);

          \node[main node] (11) [below of=1, below=1cm] {};
          \node[main node] (12) [below of=2, below=1cm] {};
          \node[main node] (13) [below of=3, below=1cm] {};
          \node[main node] (14) [below of=4, below=1cm] {};
          \node[main node] (15) [below of=5, below=1cm] {};
        
          \path[every node/.style={font=\sffamily\small}]
            (11) edge node [below=.1cm] {the} (12)
            (12) edge node [below=.1cm] {dog} (13)
            (13) edge node [below=.1cm] {barked} (14)
            (14) edge node [below=.25cm] {.} (15)
            (11) edge[bend left] node [above] {the\_dog} (13)
            (12) edge[bend left] node [above] {dog\_barked\_.} (15);

          \node[main node] (21) [below of=11, below=1cm] {};
          \node[main node] (22) [below of=12, below=1cm] {};
          \node[main node] (23) [below of=13, below=1cm] {};
          \node[main node] (24) [below of=14, below=1cm] {};
          \node[main node] (25) [below of=15, below=1cm] {};
        
%
          \path[every node/.style={font=\sffamily\small}]
            (21) edge node [below=.1cm] {the} (22)
            (22) edge node [below=.1cm] {dog} (23)
            (23) edge node [below=.1cm] {barked} (24)
            (24) edge node [below=.25cm] {.} (25)
            (21) edge[bend left] node [above] {the\_dog} (23)
            (22) edge[bend left] node [above=.3cm] {dog\_barked} (24)
            (23) edge[bend left] node [above] {barked\_.} (25);

          \node[main node] (31) [below of=21, below=.7cm] {};
          \node[main node] (32) [below of=22, below=.7cm] {};
          \node[main node] (33) [below of=23, below=.7cm] {};
          \node[main node] (34) [below of=24, below=.7cm] {};
          \node[main node] (35) [below of=25, below=.7cm] {};
        
          \path[every node/.style={font=\sffamily\small}]
            (31) edge node [above=.1cm] {the} (32)
            (32) edge[bend left] node [above=.1cm] {dog$_1$} (33)
            (33) edge[bend left] node [above=.1cm] {barked$_1$} (34)
            (34) edge node [above=.15cm] {.} (35)
            (32) edge[bend right] node [below=.1cm] {dog$_2$} (33)
            (33) edge[bend right] node [below=.1cm] {barked$_2$} (34);
            
        \node (0) [left of=1] {(a)};
        \node (10) [left of=11] {(b)};
        \node (20) [left of=21] {(c)};
        \node (30) [left of=31] {(d)};
        
        \end{tikzpicture}
    \caption{Example of (a) a single-path lattice, (b) a sparse lattice, (c) a dense lattice with $D = 2$, and (d) a multilattice with $D = 2$, for sentence ``the dog barked .''}
    \label{fig:lattice}
\end{figure}

The proposed technique is completely agnostic to the shape of the lattice, and Fig. \ref{fig:lattice} illustrates several potential varieties of lattices. Depending on how the lattice is constructed, this approach can be useful in a variety of different contexts, two of which we discuss in \S\ref{sec:instantiations}. 

\subsection{Neural Lattice Language Models}
\label{sec:hs}

There is still one missing piece in our attempt to apply neural language models to lattices.
Within our overall probability in Eq.~(\ref{eq:neuralsum}), we must calculate the probability $p(x_i^j \mid x_{<i})$ of the next segment given the history.
However, given that there are potentially an exponential number of paths through the lattice leading to $x_i$, this is not as straightforward as in the case where only one segmentation is possible.
Previous work on lattice-based language models \cite{neubig2010learning,dupont1997lattice} utilized count-based $n$-gram models, which depend on only a limited historical context at each step making it possible to compute the marginal probabilities in an exact and efficient manner through dynamic programming.
On the other hand, recurrent neural models depend on the entire context, causing them to lack this ability. Our primary technical contribution is therefore to describe several techniques for incorporating lattices into a neural framework with infinite context, by providing ways to approximate the hidden state of the recurrent neural net.

\subsubsection{Direct Approximation}
\label{subsec:da}

One approach to approximating the hidden state is the TreeLSTM framework described by \newcite{tai2015improved}.%
\footnote{This framework has been used before for calculating neural sentence representations involving lattices by \newcite{DBLP:journals/corr/SuTXL16} and \newcite{sperber2017neural}, but not for the language models that are the target of this paper.}
In the TreeLSTM formulation, new states are derived from multiple predecessors by simply summing the individual hidden and cell state vectors of each of them.
For each predecessor location $i \in A_j$, we first calculate the local hidden state $\tilde{h}$ and local cell state $\tilde{c}$ by combining the embedding $e_i^j$ with the hidden state of the LSTM at $x_{<i}$ using the standard LSTM update function as in Eq.~(\ref{eqn:lstm}):
\begin{align*}
\tilde{h}_i, \tilde{c}_i = \text{LSTM}(h_i,c_i,e_i^j,\theta) \text{ for } i \in A_j
\end{align*}

We then sum the local hidden and cell states:
\begin{align*}
\begin{split}
h_j = \sum_{i \in A_j} \tilde{h}_i
\end{split}
\begin{split}
c_j = \sum_{i \in A_j} \tilde{c}_i
\end{split}
\end{align*}

This formulation is powerful, but comes at the cost of sacrificing the probabilistic interpretation of which paths are likely.
Therefore, even if almost all of the probability mass comes through the ``true'' segmentation, the hidden state may still be heavily influenced by all of the ``bad'' segmentations as well.

\subsubsection{Monte-Carlo Approximation}
\label{subsec:af}

Another approximation that has been proposed is to sample one predecessor state from all possible predecessors, as seen in \newcite{chan2016latent}. We can calculate the total probability that we reach some prefix $x_{<j}$, and we know how much of this probability comes from each of its predecessors in the lattice, so we can construct a probability distribution over predecessors in the lattice:
\begin{align}
\label{eqn:preddist}
M(x_{<i} \mid \theta) = \frac{p(x_{<i} \mid \theta)p(x_i^j\mid x_{<i}; \theta)}{p(x_{<j} \mid \theta)}
\end{align}

Therefore, one way to update the LSTM is to sample one predecessor $x_{<i}$ from the distribution $M$ and simply set $h_j = \tilde{h}_i$ and $c_j = \tilde{c}_i$. However, sampling is unstable and difficult to train: we found that the model tended to over-sample short tokens early on during training, and thus segmented every sentence into unigrams. This is similar to the outcome reported by \newcite{chan2016latent}, who accounted for it by incorporating an $\epsilon$ encouraging exploration.

\subsubsection{Marginal Approximation}
\label{subsec:we}

In another approach, which allows us to incorporate information from all predecessors while maintaining a probabilistic interpretation, we can utilize the probability distribution $M$ to instead calculate the expected value of the hidden state:
\begin{align*}
h_j = \bm{E}_{x_{<i}\sim M} [\tilde{h_i}] = \sum_{i \in A_j} M(x_{<i} \mid \theta)\tilde{h}_i \\
c_j = \bm{E}_{x_{<i}\sim M} [\tilde{c_i}] = \sum_{i \in A_j} M(x_{<i} \mid \theta)\tilde{c}_i
\end{align*}


\subsubsection{Gumbel-Softmax Interpolation}
\label{subsec:gs}

The Gumbel-Softmax trick, or concrete distribution, described by \newcite{jang2016categorical} and \newcite{maddison2016concrete}, is a technique for incorporating discrete choices into differentiable neural computations. In this case, we can use it to select a predecessor.
The Gumbel-Softmax trick works by taking advantage of the fact that adding Gumbel noise to the pre-softmax predecessor scores and then taking the argmax is equivalent to sampling from the probability distribution.
By replacing the argmax with a softmax function scaled by a temperature $\tau$, we can get this pseudo-sampled distribution through a fully differentiable computation:
\small
\begin{align*}
N(x_{<i} \mid \theta) = \frac{\text{exp}((\text{log}(M(x_{<i} \mid \theta)) + g_i)/\tau)}{\sum_{k \in A_j} \text{exp}((\text{log}(M(x_{<k} \mid \theta)) + g_k)/\tau)} 
\end{align*}
\normalsize

This new distribution can then be used to calculate the hidden state by taking a weighted average of the states of possible predecessors:
\begin{align*}
\begin{split}
h_j = \sum_{i \in A_j}^{j-1} N(x_{<i} \mid \theta)\tilde{h}_i
\end{split}
\begin{split}
c_j = \sum_{i=j-L}^{j-1} N(x_{<i} \mid \theta)\tilde{c}_i
\end{split}
\end{align*}

When $\tau$ is large, the values of $N(x_{<i} \mid \theta)$ are flattened out; therefore, all the predecessor hidden states are summed with approximately equal weight, equivalent to the direct approximation (\S\ref{subsec:da}). On the other hand, when $\tau$ is small, the output distribution becomes extremely peaky, and one predecessor receives almost all of the weight. Each predecessor $x_{<i}$ has a chance of being selected equal to $M(x_{<i} \mid \theta)$, which makes it identical to ancestral sampling (\S\ref{subsec:af}). By slowly annealing the value of $\tau$, we can smoothly interpolate between these two approaches, and end up with a probabilistic interpretation that avoids the instability of pure sampling-based approaches.

\section{Instantiations of Neural Lattice LMs}
\label{sec:instantiations}

In this section, we introduce two instantiations of neural lattice languages models aiming to capture features of language: the existence of coherent multi-token chunks, and the existence of polysemy.

\subsection{Incorporating Multi-Token Phrases}
\label{sec:multitoken}

\subsubsection{Motivation}

Natural language phrases often demonstrate significant non-compositionality: for example, in English, the phrase ``rock and roll'' is a genre of music, but this meaning is not obtained by viewing the words in isolation. In word-level language modeling, the network is given each of these words as input, one at a time; this means it must capture the idiomaticity in its hidden states, which is quite roundabout and potentially a waste of the limited parameters in a neural network model. A straightforward solution is to have an embedding for the entire multi-token phrase, and use this to input the entire phrase to the LSTM in a single timestep. However, it is also important that the model is able to decide whether the non-compositional representation is appropriate given the context: sometimes, ``rock'' is just a rock.

Additionally, by predicting multiple tokens in a single timestep, we are able to decrease the number of timesteps across which the gradient must travel, making it easier for information to be propagated across the sentence.
This is even more useful in non-space-delimited languages such as Chinese, in which segmentation is non-trivial, but character-level modeling leads to many sentences being hundreds of tokens long.

There is also psycho-linguistic evidence which supports the fact that humans incorporate multi-token phrases into their mental lexicon. \newcite{siyanova2011adding} show that native speakers of a language have significantly reduced response time when processing idiomatic phrases, whether they are used in an idiomatic sense or not, while \newcite{bannard2008stored} show that children learning a language are better at speaking common phrases than uncommon ones.
This evidence lends credence to the idea that multi-token lexical units are a useful tool for language modeling in humans, and so may also be useful in computational models.

\subsubsection{Modeling Strategy}

The underlying lattices utilized in our multi-token phrase experiments are ``dense'' lattices: lattices where every edge (below a certain length $L$) is present (Fig. \ref{fig:lattice}, c).
This is for two reasons.
First, since every sequence of tokens is given an opportunity to be included in the path, all segmentations are candidates, which will potentially allow us to discover arbitrary types of segmentations without a prejudice towards a particular theory of which multi-token units we should be using.
Second, using a dense lattice makes minibatching very straightforward by ensuring that the computation graphs for each sentence are identical. If the lattices were not dense, the lattices of various sentences in a minibatch could be different; it then becomes necessary to either calculate a differently-shaped graph for every sentence, preventing minibatching and hurting training efficiency, or calculate and then mask out the missing edges, leading to wasted computation. Since only edges of length $L$ or less are present, the maximum in-degree of any node in the lattice $D$ is no greater than $L$, giving us the time bound O($L|X|$).

\subsubsection{Token Vocabularies}

Storing an embedding for every possible multi-token chunk would require $|V|^L$ unique embeddings, which is intractable. Therefore, we construct our multi-token embeddings by merging compositional and non-compositional representations.

\paragraph{Non-compositional Representation}

We first establish a priori a set of ``core'' chunk-level tokens that each have a dense embedding. In order to guarantee full coverage of sentences, we first add every unit-level token to this vocabulary, e.g. every word in the corpus for a word-level model. Following this, we also add the most frequent n-grams (where $1 < n \leq L$). This ensures that the vast majority of sentences will have several longer chunks appear within them, and so will be able to take advantage of tokens at larger granularities.

\paragraph{Compositional Representation}


However, the non-compositional embeddings above only account for a subset of all $n$-grams, so we additionally construct compositional embeddings for each chunk by running a BiLSTM encoder over the individual embeddings of each unit-level token within it \cite{dyer2016recurrent}. In this way, we can create a unique embedding for every sequence of unit-level tokens.

We use this composition function on chunks regardless of whether they are assigned non-compositional embeddings or not, as even high-frequency chunks may display compositional properties.
Thus, for every chunk, we compute the chunk embedding vector $x_i^j$ by concatenating the compositional embedding with the non-compositional embedding if it exists, or otherwise with an $\textless$UNK$\textgreater$ embedding.

\paragraph{Sentinel Mixture Model for Predictions}

At each timestep, we want to use our LSTM hidden state $h_t$ to assign some probability mass to every chunk with length less than $L$.
To do this, we follow \newcite{merity2016pointer} in creating a new ``sentinel'' token $\textit{\textless s\textgreater}$ and adding it to our vocabulary. At each timestep, we first use our neural network to calculate a score for each chunk $C$ in our vocabulary, including the sentinel token. We do a softmax across these scores to assign a probability $p_{\textit{main}}(C_{t+1} \mid h_t; \theta)$ to every chunk in our vocabulary, and also to $\textit{\textless s\textgreater}$. For token sequences not represented in our chunk vocabulary, this probability $p_{\textit{main}}(C_{t+1} \mid h_t; \theta) = 0$.

Next, the probability mass assigned to the sentinel value, $p_{\textit{main}}(\textit{\textless s\textgreater} \mid h_t; \theta)$, is distributed across all possible tokens sequences of length less than $L$, using another LSTM with parameters $\theta_{\textit{sub}}$. Similar to \newcite{jozefowicz2016exploring}, this sub-LSTM is initialized by passing in the hidden state of the main lattice LSTM at that timestep. This gives us a probability for each sequence $p_{\textit{sub}}(c_1,c_2,\ldots,c_L \mid h_t; \theta_\textit{sub})$.

The final formula for calculating the probability mass assigned to a specific chunk $C$ is:
\begin{align*}
p(C \mid h_t; \theta) = &p_{\textit{main}}(C \mid h_t; \theta) + \\
                        &p_{\textit{main}}(\textit{\textless s\textgreater} \mid h_t; \theta) p_{\textit{sub}}(C \mid h_t; \theta_\textit{sub}) \nonumber
\end{align*}

\subsection{Incorporating Polysemous Tokens}
\label{sec:polysemy}

\subsubsection{Motivation}

A second shortcoming of current language modeling approaches is that each word is associated with only one embedding.
For highly polysemous words, a single embedding may be unable to represent all meanings effectively.

There has been past work in word embeddings which has shown that using multiple embeddings for each word is helpful in constructing a useful representation. \newcite{athiwaratkun2017multimodal} represented each word with a multimodal Gaussian distribution and demonstrated that embeddings of this form were able to outperform more standard skip-gram embeddings on word similarity and entailment tasks. Similarly, \newcite{DBLP:journals/corr/ChenQJH15} incorporate standard skip-gram training into a Gaussian mixture framework and show that this improves performance on several word similarity benchmarks.

When a polysemous word is represented using only a single embedding in a language modeling task, the multimodal nature of the true embedding distribution may causes the resulting embedding to be both high-variance and skewed from the positions of each of the true modes. Thus, it is likely useful to represent each token with multiple embeddings when doing language modeling.

\subsubsection{Modeling Strategy}

For our polysemy experiments, the underlying lattices are ``multilattices'': lattices which are also multigraphs, and can have any number of edges between any given pair of nodes (Fig. \ref{fig:lattice}, d). Lattices set up in this manner allow us to incorporate multiple embeddings for each word. Within a single sentence, any pair of nodes corresponds to the start and end of a particular subsequence of the full sentence, and is thus associated with a specific token; each edge between them is a unique embedding for that token.
While many strategies for choosing the number of embeddings exist in the literature \cite{neelakantan2014efficient}, in this work, we choose a number of embeddings $E$ and assign that many embeddings to each word. This ensures that the maximum in-degree of any node in the lattice $D$, is no greater than $E$, giving us the time bound O($E|X|$).

In this work, we do not explore models that include both chunk vocabularies and multiple embeddings. However, combining these two techniques, as well as exploring other, more complex lattice structures, is an interesting avenue for future work.

\section{Experiments}
\label{sec:experiments}

\subsection{Data}

We perform experiments on two languages: English and Chinese, which provide an interesting contrast in linguistic features.%
\footnote{Code to reproduce datasets and experiments is available at: \url{http://github.com/jbuckman/neural-lattice-language-models}}

In English, the most common benchmark for language modeling recently is the Penn Treebank, specifically the version preprocessed by \newcite{mikolovptb}.
However, this corpus is limited by being relatively small, only containing approximately 45,000 sentences, which we found to be insufficient to effectively train lattice language models.%
\footnote{Experiments using multi-word units resulted in overfitting, regardless of normalization and hyperparameter settings.}
Thus, we instead used the Billion Word Corpus \cite{chelba2013one}.
Past experiments on the BWC typically modeled every word without restricting the vocabulary, which results in a number of challenges regarding the modeling of open vocabularies that are orthogonal to this work. Thus, we create a preprocessed version of the data in the same manner as Mikolov, lowercasing the words, replacing numbers with $<$N$>$ tokens, and $<$UNK$>$ing all words beyond the ten thousand most common. Additionally, we restricted the data set to only include sentences of length 50 or less, ensuring that large minibatches could fit in GPU memory. Our subsampled English corpus contained 29,869,166 sentences, of which 29,276,669 were used for training, 5,000 for validation, and 587,497 for testing. To validate that our methods scale up to larger language modeling scenarios, we also report a smaller set of large-scale experiments on the full billion word benchmark in Appendix A.

In Chinese, we ran experiments on a subset of the Chinese GigaWord corpus.
Chinese is also particularly interesting because unlike English, it does not use spaces to delimit words, so segmentation is non-trivial.
Therefore, we used a character-level language model for the baseline, and our lattice was composed of multi-character chunks. We used sentences from \textit{Guangming Daily}, again $<$UNK$>$ing all but the 10,000 most common tokens and restricting the selected sentences to only include sentences of length 150 or less. Our subsampled Chinese corpus included 934,101 sentences for training, 5,000 for validation, and 30,547 for testing.


\subsection{Main Experiments}

We compare a baseline LSTM model, dense lattices of size 1, 2, and 3, and a multilattice with 2 and 3 embeddings per word. 

The implementation of our networks was done in DyNet \cite{neubig2017dynet}.%
 All LSTMs had 2 layers, each with hidden dimension of 200. Variational dropout \cite{gal2016theoretically} of .2 was used on the Chinese experiments, but hurt performance on the English data, so it was not used. The 10,000 word embeddings each had dimension 256. For lattice models, chunk vocabularies were selected by taking the 10,000 words in the vocabulary and adding the most common 10,000 $n$-grams with $1 < n \leq L$. The weights on the final layer of the network were tied with the input embeddings, as done by \cite{press2016using,inan2016tying}. In all lattice models, hidden states were computed using weighted expectation (\S\ref{subsec:we}) unless mentioned otherwise. In multi-embedding models, embedding sizes were decreased so as to maintain the same total number of parameters. All models were trained using the Adam optimizer with a learning rate of .01 on a NVIDIA K80 GPU.
The results can be seen in Table \ref{lm-results-en} and Table \ref{lm-results-zh}.

In the multi-token phrase experiments, many additional parameters are accrued by the BiLSTM encoder and sub-LSTM predictive model, making them not strictly comparable to the baseline. To account for this, we include results for $L=1$, which, like the baseline LSTM approach, fails to leverage multi-token phrases, but includes the same number of parameters as $L=2$ and $L=3$.

In both the English and Chinese experiments, we see the same trend: increasing the maximum lattice size decreases the perplexity, and for $L=2$ and above, the neural lattice language model outperforms the baseline. Similarly, increasing the number of embeddings per word decreases the perplexity, and for $E=2$ and above, the multiple-embedding model outperforms the baseline.

\begin{table}[]
\centering
\caption{Results on English language modeling task}
\label{lm-results-en}
\begin{tabular}{|c|c|c|}
\hline
Model             & Valid. Perp.     & Test Perp.  \\ \hline \hline
Baseline          &  47.64           & 48.62       \\ \hline \hline
Multi-Token ($L=1$)   &  45.69           & 47.21       \\ \hline
Multi-Token ($L=2$)   &  44.15           & 46.12       \\ \hline
Multi-Token ($L=3$)   &  45.19           & 46.84       \\ \hline \hline
Multi-Emb ($E=2$)        &  44.80           & 46.32       \\ \hline
Multi-Emb ($E=3$)        &  \textbf{42.76}  &  \textbf{43.78}    \\ \hline
\end{tabular}
\end{table}

\begin{table}[]
\centering
\caption{Results on Chinese language modeling task}
\label{lm-results-zh}
\begin{tabular}{|c|c|c|}
\hline
Model             & Valid. Perp.      & Test Perp.  \\ \hline \hline
Baseline          &  41.46            & 40.72       \\ \hline \hline
Multi-Token ($L=1$)   &  49.86            & 50.99       \\ \hline
Multi-Token ($L=2$)   &  38.61            & 37.22       \\ \hline
Multi-Token ($L=3$)   &  \textbf{33.01}   & \textbf{32.19}       \\ \hline \hline
Multi-Emb ($E=2$) &  40.30            & 39.28       \\ \hline
Multi-Emb ($E=3$) &  45.72            & 44.40       \\ \hline
\end{tabular}
\end{table}

\subsection{Hidden State Calculation Experiments}

We compare the various hidden-state calculation approaches discussed in Section \ref{sec:hs} on the English data using a lattice of size $L=2$ and dropout of .2. These results can be seen in Table \ref{hs-results}.

For all hidden state calculation techniques, the neural lattice language models outperform the LSTM baseline. The ancestral sampling technique used by \newcite{chan2016latent} is worse than the others, which we found to be due to it getting stuck in a local minimum which represents almost everything as unigrams. There is only a small difference between the perplexities of the other techniques.

\begin{figure*}[t!]
    \centering
    \includegraphics[scale=.5]{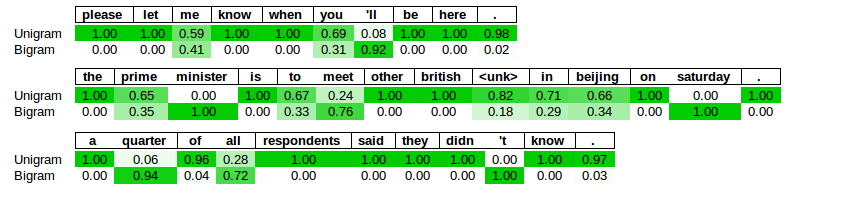}
    \caption{Segmentation of three sentences randomly sampled from the test corpus, using $L=2$. Green numbers show probability assigned to token sizes. For example, the first three words in the first sentence have a 59\% and 41\% chance of being ``please let me'' or ``please let\_me'' respectively. Boxes around words show greedy segmentation.}
    \label{fig:segmentations}
\end{figure*}

\begin{table}[]
\centering
\caption{Hidden state calculation comparison results}
\label{hs-results}
\small
\begin{tabular}{|c|c|c|}
\hline
Model                & Valid. Perp.           & Test Perp.       \\ \hline \hline
Baseline              &  64.18                & 60.67           \\ \hline
Direct (\S\ref{subsec:da}) &  59.74                & 55.98           \\ \hline
Monte Carlo (\S\ref{subsec:af}) &  62.97                & 59.08           \\ \hline
Marginalization (\S\ref{subsec:we}) &  \textbf{58.62}                & \textbf{55.06}           \\ \hline
GS Interpolation (\S\ref{subsec:gs}) &  59.19                & 55.73           \\ \hline
\end{tabular}
\end{table}

\subsection{Discussion and Analysis}
\label{sec:discussion}

Neural lattice language models convincingly outperform an LSTM baseline on the task of language modeling.
One interesting note is that in English, which is already tokenized into words and highly polysemous, utilizing multiple embeddings per word is more effective than including multi-word tokens.
In contrast, in the experiments on the Chinese data, increasing the lattice size of the multi-character tokens is more important than increasing the number of embeddings per character. 
This corresponds to our intuition; since Chinese is not tokenized to begin with, utilizing models that incorporate segmentation and compositionality of elementary units is very important for effective language modeling.

\begin{table*}[t!]
\centering
\caption{Comparison of randomly-selected contexts of several words selected from the vocabulary of the Billion Word Corpus, in which the model preferred one embedding over the other.}
\label{multi-viz}
\begin{tabular}{|l||l|}

\hline
\textbf{rock$_1$} & \textbf{rock$_2$} \\ \hline
...at the $<$unk$>$ pop , \textit{rock} and jazz... & ...including hsbc , northern \textit{rock} and...                       \\ \hline
...a little bit $<$unk$>$ \textit{rock} ,...        & ...pakistan has a $<$unk$>$ \textit{rock} music scene...          \\ \hline
...on light \textit{rock} and $<$unk$>$ stations... & ...spokesman for round \textit{rock} , $<$unk$>$... \\ \hline \hline

\textbf{bank$_1$} & \textbf{bank$_2$} \\ \hline
...being a \textit{bank} holiday in... & ...the \textit{bank} of england has...                       \\ \hline
...all the us \textit{bank} runs and...        & ...with the royal \textit{bank} of scotland...          \\ \hline
...by getting the \textit{bank} 's interests... & ...development \textit{bank} of japan and the... \\ \hline \hline

\textbf{page$_1$} & \textbf{page$_2$} \\ \hline
...on \textit{page} $<$unk$>$ of the... & ...was it front \textit{page} news...                       \\ \hline
...a source told \textit{page} six ....        & ...himself , tony \textit{page} , the former ...          \\ \hline
...on \textit{page} $<$unk$>$ of the... & ...sections of the \textit{page} that discuss... \\ \hline \hline

\textbf{profile$_1$} & \textbf{profile$_2$} \\ \hline
...( $<$unk$>$ : quote , \textit{profile} , research )... & ...so $<$unk$>$ the \textit{profile} of the city...                       \\ \hline
...( $<$unk$>$ : quote , \textit{profile} , research )...        & ...the highest \textit{profile} $<$unk$>$ held by...          \\ \hline
...( $<$unk$>$ : quote , \textit{profile} , research )... & ...from high i , elite schools ,... \\ \hline \hline

\textbf{edition$_1$} & \textbf{edition$_2$} \\ \hline
... of the second \textit{edition} of windows... & ...of the new york \textit{edition} . ...                       \\ \hline
... this month 's \textit{edition} of$<$unk$>$ , the ...        & ...of the new york \textit{edition} . ...          \\ \hline
...forthcoming d.c. \textit{edition} of the hit... & ...of the new york \textit{edition} . ... \\ \hline \hline

\textbf{rodham$_1$} & \textbf{rodham$_2$} \\ \hline
...senators hillary \textit{rodham} clinton and... &                     \\ \hline
...making hillary \textit{rodham} clinton his...        &       \\ \hline
...hillary \textit{rodham} clinton 's campaign has... &   \\ \hline

\end{tabular}
\end{table*}

To calculate the probability of a sentence, the neural lattice language model implicitly marginalizes across latent segmentations. By inspecting the probabilities assigned to various edges of the lattice, we can visualize these segmentations, as is done in Fig. \ref{fig:segmentations}. The model successfully identifies bigrams which correspond to non-compositional compounds, like ``prime minister'', and bigrams which correspond to compositional compounds, such as ``a quarter''. Interestingly, this does not occur for all high-frequency bigrams; it ignores those that are not inherently meaningful, such as ``$<$UNK$>$ in'', yielding qualitatively good phrases.

In the multiple-embedding experiments, it is possible to see which of the two embeddings of a word was assigned the higher probability for any specific test-set sentence. In order to visualize what types of meanings are assigned to each embedding, we select sentences in which one embedding is preferred, and look at the context in which the word is used. Several examples of this can be seen in Table \ref{multi-viz}; it is clear from looking at these examples that the system does learn distinct embeddings for different senses of the word. What is interesting, however, is that it does not necessarily learn intuitive semantic meanings; instead it tends to group the words by the context in which they appear. In some cases, like \textbf{profile} and \textbf{edition}, one of the two embeddings simply captures an idiosyncrasy of the training data.

Additionally, for some words, such as \textbf{rodham} in Table \ref{multi-viz}, the system always prefers one embedding. This is promising, because it means that in future work it may be possible to further improve accuracy and training efficiency by assigning more embeddings to polysemous words, instead of assigning the same number of embeddings to all words.

\section{Related Work}
\label{related}

Past work that utilized lattices in neural models for natural language processing centers around using these lattices in the encoder portion of machine translation. \newcite{DBLP:journals/corr/SuTXL16} utilized a variation of the Gated Recurrent Unit that operated over lattices, and preprocessed lattices over Chinese characters that allowed it to effectively encode multiple segmentations. Additionally, \newcite{sperber2017neural} proposed a variation of the TreeLSTM with the goal of creating an encoder over speech lattices in speech-to-text.
Our work tackles language modeling rather than encoding, and thus addresses the issue of marginalization over the lattice.

Another recent work which marginalized over multiple paths through a sentence is \newcite{ling2016latent}. The authors tackle the problem of code generation, where some components of the code can be copied from the input, via a neural network.
Our work expands on this by handling multi-word tokens as input to the neural network, rather than passing in one token at a time.

Neural lattice language models improve accuracy by helping the gradient flow over smaller paths, preventing vanishing gradients. Many hierarchical neural language models have been proposed with a similar objective \newcite{koutnik2014clockwork,zhou2017chunk}.  Our work is distinguished from these by the use of latent token-level segmentations that capture meaning directly, rather than simply being high-level mechanisms to encourage gradient flow.

\newcite{chan2016latent} propose a model for predicting characters at multiple granularities in the decoder segment of a machine translation system. Our work expands on theirs by considering the entire lattice at once, rather than considering a only a single path through the lattice via ancestral sampling. This allows us to train end-to-end without the model collapsing to a local minimum, with no exploration bonus needed. Additionally, we propose a more broad class of models, including those incorporating polysemous words, and apply our model to the task of word-level language modeling, rather than character-level transcription.

Concurrently to this work, \newcite{van2017multiscale} have proposed a neural language model that can similarly handle multiple scales.
Our work is differentiated in that it is more general: utilizing an open multi-token vocabulary, proposing multiple techniques for hidden state calculation, and handling polysemy using multi-embedding lattices.

\section{Future Work}

In the future, we would like to experiment with utilizing neural lattice language models in extrinsic evaluation, such as machine translation and speech recognition. Additionally, in the current model, the non-compositional embeddings must be selected a priori, and may be suboptimal. We are exploring techniques to store fixed embeddings dynamically, so that the non-compositional phrases can be selected as part of the end-to-end training.

\section{Conclusion}

In this work, we have introduced the idea of a neural lattice language model, which allows us to marginalize over all segmentations of a sentence in an end-to-end fashion. In our experiments on the Billion Word Corpus and Chinese GigaWord corpus, we demonstrated that the neural lattice language model beats an LSTM-based baseline at the task of language modeling, both when it is used to incorporate multiple-word phrases and multiple-embedding words.
Qualitatively, we observed that the latent segmentations generated by the model correspond well to human intuition about multi-word phrases, and that the varying usage of words with multiple embeddings seems to also be sensible.

\bibliographystyle{acl2012}

\appendix
\section{Large-Scale Experiments}

To verify that our findings scale to state-of-the-art language models, we also compared a baseline model, dense lattices of size 1 and 2, and a multilattice with 2 embeddings per word on the full byte-pair encoded Billion Word Corpus.

In this set of experiments, we take the full Billion Word Corpus, and apply byte-pair encoding as described by \newcite{sennrich2015neural} to construct a vocabulary of 10,000 sub-word tokens. Our model consists of three LSTM layers, each with 1500 hidden units. We train the model for a single epoch over the corpus, using the Adam optimizer with learning rate .0001 on a P100 GPU. We use a batch size of 40, and variational dropout of 0.1. The 10,000 sub-word embeddings each had dimension 600. For lattice models, chunk vocabularies were selected by taking the 10,000 sub-words in the vocabulary and adding the most common 10,000 $n$-grams with $1 < n \leq L$. The weights on the final layer of the network were tied with the input embeddings, as done by \newcite{press2016using,inan2016tying}. In all lattice models, hidden states were computed using weighted expectation (\S\ref{subsec:we}). In multi-embedding models, embedding sizes were decreased so as to maintain the same total number of parameters.

Results of these experiments are in Table \ref{lm-results-big}. The performance of the baseline model is roughly on par with that of state-of-the-art models on this database; differences can be explained by model size and hyperparameter tuning. The results show the same trend as the results of our main experiments, indicating that the performance gains shown by our smaller neural lattice language models generalize to the much larger datasets used in state-of-the-art systems.

\begin{table}[]
\centering
\caption{Results on large-scale Billion Word Corpus}
\small
\label{lm-results-big}
\begin{tabular}{|c|c|c|c|}
\hline
Model                 & Valid. & Test  & Sec./ \\
                      & Perp.  & Perp. & Batch \\ \hline \hline
Baseline              & 54.1             & 37.7             & .45          \\ \hline \hline
Multi-Token ($L=1$)   & 54.2             & 37.4             & .82           \\ \hline
Multi-Token ($L=2$)   & 53.9             & 36.4             & 4.85           \\ \hline
Multi-Emb ($E=2$)     & \textbf{53.8}    & \textbf{35.2}    & 2.53           \\ \hline
\end{tabular}
\end{table}

\section{Chunk Vocabulary Size}

\begin{table}[]
\centering
\caption{Vocabulary size comparison}
\label{size-results}
\begin{tabular}{|c|c|c|}
\hline
Model             & Valid. Perp. & Test Perp. \\ \hline
Baseline          & 64.18                 & 60.67           \\ \hline
10000-chunk vocab & 58.62                 & 55.06           \\ \hline
20000-chunk vocab & \textbf{57.40}                 & \textbf{54.15}           \\ \hline
\end{tabular}
\end{table}

We compare a 2-lattice with a non-compositional chunk vocabulary of 10,000 phrases with a 2-lattice with a non-compositional chunk vocabulary of 20,000 phrases. The results can be seen in Table \ref{size-results}. Doubling the number of non-compositional embeddings present decreases the perplexity, but only by a small amount. This is perhaps to be expected, given that doubling the number of embeddings corresponds to a large increase in the number of model parameters for phrases that may have less data with which to train them.

\end{document}